\documentclass{article}

\usepackage[nonatbib, preprint]{nips_2018}

\usepackage{times}
\usepackage{cite}
\usepackage{subfigure}
\usepackage{graphicx}
\usepackage{amsmath}
\usepackage{bm}
\usepackage{url}
\usepackage{stmaryrd}
\usepackage{balance}
\usepackage{amssymb}
\usepackage{pgfplots}
\usepackage{pifont}
\usepackage[usenames,dvipsnames]{pstricks}
\usepackage{epsfig}
\usepackage{booktabs}
\usepackage{pgffor}
\usepackage{tikz}
\usepackage{multirow}

\newcommand{\mb}{\mathbf}

\newcommand{\our}{\textsc{Gen}}

\newcommand{\linemodel}{\textsc{LINE}}
\newcommand{\deepwalk}{\textsc{DeepWalk}}
\newcommand{\nodevec}{\textsc{node2vec}}
\newcommand{\hpe}{\textsc{HPE}}

\usepackage{algorithm}
\usepackage{algorithmic}

\begin{document}
\title{GEN Model: An Alternative Approach to Deep Neural Network Models}



\author{Jiawei~Zhang$^\star$, Limeng Cui$^\dagger$, Fisher B. Gouza$^\dagger$\\
$^\star$IFM Lab, Florida State University, FL, USA\\
$^\dagger$University of Chinese Academy of Sciences, Beijing, China\\
jzhang@cs.fsu.edu, lmcui932@163.com, fisherbgouza@gmail.com}

\maketitle


\vspace{-25pt}

\begin{abstract}

\vspace{-10pt}

In this paper, we introduce an alternative approach, namely {\our} (Genetic Evolution Network) Model, to the deep learning models. Instead of building one single deep model, {\our} adopts a genetic-evolutionary learning strategy to build a group of unit models generations by generations. Significantly different from the well-known representation learning models with extremely deep structures, the unit models covered in {\our} are of a much shallower architecture. In the training process, from each generation, a subset of unit models will be selected based on their performance to evolve and generate the child models in the next generation. {\our} has significant advantages compared with existing deep representation learning models in terms of both learning effectiveness, efficiency and interpretability of the learning process and learned results. Extensive experiments have been done on diverse benchmark datasets, and the experimental results have demonstrated the outstanding performance of {\our} compared with the state-of-the-art baseline methods in both effectiveness of efficiency.

\end{abstract}

\vspace{-20pt}
\section{Introduction}\label{sec:intro}
\vspace{-10pt}

In recent years, deep learning \cite{GBC16, LBH15}, a rebranding of deep neural network and other related research works, has achieved a great impact and success in various applications. With multiple hidden layers, the deep learning models have the capacity to capture the projections from the input space (of data) to the output space, whose outstanding performance have been widely illustrated in network embedding \cite{WCZ16, CHTQAH15, PAS14}, text mining \cite{ASKR12, MH09, DHK13, HDYDMJSVNSK12}, and computer vision \cite{KSH12, LBH15, KTSLSL14}.

Meanwhile, deep learning models also suffer from serious criticism due to their disadvantages in the demands of (1) a large amount of training data, (2) powerful computational facilities, (3) heavy parameter tuning efforts, and the lack of (4) theoretic explanation of the learning results. These disadvantages greatly hinder the application of deep learning in many areas that cannot provide such requirements. Due to these reasons, deep learning research and application works are mostly carried out within/via the collaboration with several big technical companies, but the models proposed by them (involving hundreds of hidden layers, billions of parameters, and large server clusters \cite{Dean:2012:LSD:2999134.2999271}) can hardly be applied in other real-world applications.

In this paper, we propose the {\our} (Genetic Evolution Network) model, which can work as an alternative approach to the deep learning models. Instead of building one single deep model, {\our} adopts a genetic-evolution learning strategy instead, which learns a group of unit models generations by generations. Here, the unit models can be either traditional ``shallow'' machine learning models or the deep models with a relatively ``shallower'' and ``narrower'' structure. Each unit model will be trained with a batch of small-sized training instances sampled form the dataset. By selecting the good unit models from each generation (according to their performance on a validation set), {\our} will evolve and generate the next generation of unit modes with the probabilistic genetic algorithm, where the selection and crossover probabilities from parent models to the child model are highly dependent on their evaluated fitness scores. Finally, the learning results of each instance will be combined from each unit model via effective ensemble learning techniques.

From the bionics perspective, {\our} effectively models the evolution of creatures from generations to generations. Each unit network model in a generation can be treated as a unit creature, which will receive training from the external world (i.e., training data). Different creatures have different experiences, and will get a different subsets of training data, which leads to different connection weights among their brain neurons. For creatures whose brain neuron connections suitable for the environment (i.e., validation set), they will have a larger chance to survive and generate their offsprings; while the parent creature achieving better performance will have a greater chance to inherit their brain connection weights to the child models. 

From the computation perspective, the unit models in each generation of {\our} are of a much simpler architecture, learning of which can be accomplished with much less training data, computational resources and hyper-parameter tuning efforts. In addition, the relatively ``shallower'' structure of unit models will also significantly enhance the interpretability of the unit models themselves as well as the learning results. Furthermore, the sound theoretical foundations about genetic algorithm \cite{M98} and ensemble learning \cite{D00} will also help explain the information inheritation through generations  and output results integration from the unit models.

In this paper, we will use network embedding problem \cite{WCZ16, CHTQAH15, PAS14} (applying autoencoder as the unit model) as an example to illustrate the {\our} model. Meanwhile, applications of {\our} on other data categories (e.g., images and raw feature inputs) with CNN and MLP as the unit model will be provided in Section~\ref{subsec:other_results}. This paper is organized as follows. Some related works will be talked about in Section~\ref{sec:relatedwork}. Model {\our} will be introduced in Section~\ref{sec:method}, whose performance will be analyzed in Section~\ref{sec:analysis} and evaluated in Section~\ref{sec:experiment}. Finally, we will conclude this paper in Section~\ref{sec:conclusion}.
\vspace{-10pt}
\section{Related Works} \label{sec:relatedwork}
\vspace{-10pt}

The essence of deep learning is to compute hierarchical features or representations of the observational data \cite{GBC16, LBH15}. With the surge of deep learning research and applications in recent years, lots of research works have appeared to apply the deep learning methods, like deep belief network \cite{HOT06}, deep Boltzmann machine \cite{SH09}, deep neural network \cite{J02, KSH12} and deep autoencoder model \cite{VLLBM10}, in various applications, like speech and audio processing \cite{DHK13, HDYDMJSVNSK12}, language modeling and processing \cite{ASKR12, MH09}, information retrieval \cite{H12, SH09}, objective recognition and computer vision \cite{LBH15}, as well as multimodal and multi-task learning \cite{WBU10, WBU11}. Traditional deep learning models have too many disadvantages, Zhou et al. introduce the deep forest as an alternative approach in \cite{ijcai2017-497}.


In recent years, many research works propose to embed network data into a low-dimensional feature space, i.e., the network embedding problem, in which nodes are represented as feature vectors. In graphs, the relation can be treated as a translation of the entities, and many translation based embedding models have been proposed, like TransE \cite{BUGWY13}, TransH \cite{WZFC14} and TransR \cite{LLSLZ15}. At the same time, many network embedding works based on random walk model and deep learning models have been introduced, like Deepwalk \cite{PAS14}, LINE \cite{TQWZYM15}, node2vec \cite{GL16}, HNE \cite{CHTQAH15} and DNE \cite{WCZ16}. Perozzi et al. extends the word2vec model \cite{MSCCD13} to the network scenario and introduce the Deepwalk algorithm \cite{PAS14}. Tang et al. \cite{TQWZYM15} propose to embed the networks with LINE algorithm, which can preserve both the local and global network structures. Grover et al. \cite{GL16} introduce a flexible notion of a node's network neighborhood and design a biased random walk procedure to sample the neighbors.


\vspace{-10pt}
\section{Proposed Methods}\label{sec:method}
\vspace{-10pt}

In this section, we will introduce the detailed information about the {\our} model, including its overall architecture, the unit model initialization, training, evolution and the final result ensemble methods.

\vspace{-10pt}
\subsection{Overall Model Architecture}
\vspace{-8pt}

\begin{figure*}[!t]
\vspace{-30pt}
 \centering    
 \begin{minipage}[l]{0.9\columnwidth}
  \centering
    \includegraphics[width=1.0\textwidth]{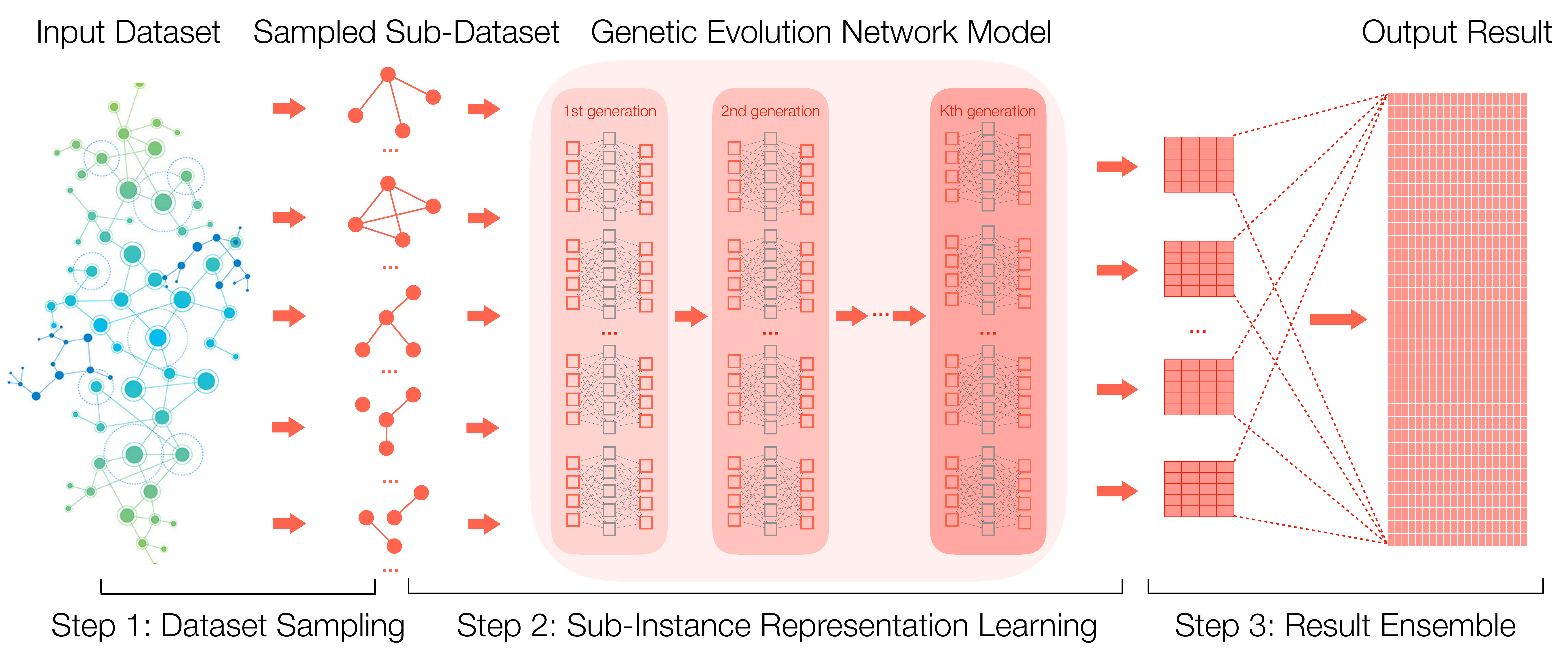}
 \end{minipage}
 \vspace{-8pt}
\caption{Overall Architecture of {\our} Model.}\label{fig:framework}\vspace{-20pt}
\end{figure*}

The overall architecture of {\our} is shown in Figure~\ref{fig:framework}, given the input dataset, {\our} learns the output results with three main steps. In Figure~\ref{fig:framework}, we use the network structured data as an example to illustrate how the {\our} model works.

\vspace{-4pt}

$\bullet$ \textit{Step 1. Dataset Sampling}: Based on the input dataset, the \textit{dataset sampling} step aims at sampling a set of small-sized data instance to compose the sub-instance pool. Instead of using the original large-scale input data, the model training and validation in the following steps will all be based on the sampled sub-instance pool.

\vspace{-4pt}

$\bullet$ \textit{Step 2. Model Learning}: The {\our} model involves multiple generations of unit models, starting from the $1_{st}$-generation to the $K_{th}$-generation as shown in Figure~\ref{fig:framework}. For the unit models in each generation, they will be trained with the traditional stochastic gradient descent method. These trained unit models will be further evolved to the new generations via the \textit{fitness evaluation}, \textit{crossover}, \textit{mutation} and \textit{selection} operations to be introduced as follows respectively.

\vspace{-4pt}

$\bullet$ \textit{Step 3. Output Result Ensemble}: For all the data instances in the pool, each unit model in the $K_{th}$ generation of {\our} can learns an output, which will be effectively integrated together to generate the final output result.

\vspace{-4pt}

Detailed Information about these three steps will be introduced in the following subsections.


\vspace{-10pt}
\subsection{Sub-Instance Pool Sampling}
\vspace{-8pt}

Depending on the input data categories, different types of sampling strategies can be adopted. For instance, for the network structured data as shown in Figure~\ref{fig:framework}, random (biased or unbiased) node/edge sampling \cite{ANK13}, and the BFS (breadth first search)/DFS (depth first search) based sampling strategies can all be adopted. Let $G = (\mathcal{V}, \mathcal{E})$ denote the input large-scale network data with the node set $\mathcal{V}$ and edge set $\mathcal{E}$ respectively. Formally, we can represent the sampled sub-networks from $G$ as set $\mathcal{P} = \{g_1, g_2, \cdots, g_n\}$, where each sampled sub-network instance can be represented as $g = (\mathcal{V}_g, \mathcal{E}_g)$ with $\mathcal{V}_g \subset \mathcal{V}$ and $\mathcal{E}_g \subset \mathcal{E}$ denoting its node and edge sets respectively. These sampled sub-networks in $\mathcal{P}$ are much smaller than the original input network $G$ in terms of both node and edge numbers. For the other types of data, e.g., images and text data, a similar sampled pool of sub-instances can be obtained with different sampling strategies. With the small-sized sub-instances in pool $\mathcal{P}$, the learning process of the unit models will be much more efficient.

\vspace{-10pt}
\subsection{Autoencoder Model Description}
\vspace{-8pt}

Depending on the input data and problem settings, different learning models can be used as the unit model in composing {\our}. In this part, we will provide a brief introduction to the autoencoder model, an extension model from which will be used as the unit model for representation learning of networked data in {\our} in this paper.

Autoencoder \cite{VLLBM10} is an unsupervised neural network model, involving two steps: encoder and decoder. The encoder step projects the original feature vector to a latent feature space, while the decoder step recovers the latent feature representation to a reconstruction space. Formally, let $\mb{x}_i$ represent the feature vector of instance $i$, and $\mb{y}^1_i, \mb{y}^2_i, \cdots, \mb{y}^o_i$ be the corresponding latent feature representations at hidden layers $1, 2, \cdots, o$ in the encoder step (where $o=1$ for the model used in {\our} in this paper). The encoding result in the objective feature space can be denoted as $\mb{z}_i \in \mathbb{R}^{d}$ of dimension $d$. In the decoder step, based on vector $\mb{z}_i$, it outputs a reconstructed vector $\hat{\mb{x}}_i$ (of the same dimension as $\mb{x}_i$). The latent feature vectors in the decoder can be represented as $\hat{\mb{y}}^{o}_i, \hat{\mb{y}}^{o-1}_i, \cdots, \hat{\mb{y}}^{1}_i$. The relationship among these variables can be represented with the following equations:\vspace{-6pt}
\begin{alignat}{2}
\hspace{-2pt}
\begin{cases}
& \hspace{-9pt} \mbox{\# Encoder: } \\
& \hspace{-9pt} \mb{y}^1_i = \sigma (\mb{W}^1 \mb{x}_i + \mb{b}^1),\\
& \hspace{-9pt} \mb{y}^k_i = \sigma (\mb{W}^k \mb{y}^{k-1}_i + \mb{b}^k),\forall k \in \{2,\cdots, o\},\\
& \hspace{-9pt} \mb{z}_i = \sigma (\mb{W}^{o+1} \mb{y}^o_i + \mb{b}^{o+1}).
\end{cases}
\hspace{-4pt}
\begin{cases}
& \hspace{-9pt} \mbox{\# Decoder: }\\
& \hspace{-9pt} \hat{\mb{y}}^o_i = \sigma (\hat{\mb{W}}^{o+1} \mb{z}_i + \hat{\mb{b}}^{o+1}),\\
& \hspace{-9pt} \hat{\mb{y}}^{k-1}_i = \sigma (\hat{\mb{W}}^k \hat{\mb{y}}^{k}_i + \hat{\mb{b}}^k), \forall k \in \{2,\cdots, o\},\\
& \hspace{-9pt} \hat{\mb{x}}_i = \sigma(\hat{\mb{W}}^1 \hat{\mb{y}}^{1}_i + \hat{\mb{b}}^1),
\end{cases}
\end{alignat}

\vspace{-10pt}
\noindent where $\mb{W}$, $\mb{b}$ (and $\hat{\mb{W}}$, $\hat{\mb{b}}$) with different superscripts denotes the weights and biases variables in the encoder (and decoder) step of the autoencoder model respectively.

The objective of autoencoder is to minimize the loss between the input feature vector $\mb{x}_i$ and the reconstructed feature vector $\hat{\mb{x}}_i$ of data instances. Formally, the loss term can be represented as\vspace{-5pt}
\begin{equation}
\mathcal{L}_e = \sum_{i} \left \| {\mb{x}}_i - \hat{\mb{x}}_i\right\|_2^2.\end{equation}
\vspace{-10pt}

\vspace{-10pt}
\subsection{Generation Population Initialization}
\vspace{-8pt}

Model {\our} is not a static model, which keeps evolving, where \textit{generation} works as the evolution unit. Formally, we can represent these generations as set $\{\mathcal{G}^{(1)}, \mathcal{G}^{(2)}, \cdots, \mathcal{G}^{(K)}\}$, where $\mathcal{G}^{(1)}$ is the initial generation and $\mathcal{G}^{(K)}$ is the final generation. 

Formally, each generation in {\our} involves $m$ unit models, where the initial genneration can be denoted as $\mathcal{G}^{(1)} = \{M^{(1)}_1, M^{(1)}_2, \cdots, M^{(1)}_m\}$ and $M^{(1)}_i, \forall i \in \{1, 2, \cdots, m\}$ denotes the unit model. For model $M^{(1)}_i$, its variables to be learned can be denoted as vector $\mb{\theta}^{(1)}_i$. For the unit models in the initial generation, their parameters are initialized with random values; while the parameters for unit models in later generations will be inherited from their parent models instead.

\vspace{-10pt}
\subsection{Generation Training}
\vspace{-8pt}

For each unit model $M^{(j)}_i \in \mathcal{G}^{(j)}$, a training batch will be sampled for it from the training pool $\mathcal{P}$ introduced before, i.e., $\mathcal{T}^{(j)}_i \subset \mathcal{P}$. Formally, for sub-network $g \in \mathcal{T}^{(j)}_i$, we can represent its network structure as an adjacency matrix $\mb{A} \in \{0, 1\}^{|\mathcal{V}_g| \times |\mathcal{V}_g|}$, where the rows denote the neighborhood features of nodes in the network. According to the introduction of autoencoder model, we can define representation learning loss of network $g$ as \vspace{-8pt}
\begin{align}
\mathcal{L}_e(g) &= \sum_{v_i \in \mathcal{V}_g} \left\| \mb{A}(i,:) - \hat{\mb{A}}(i,:) \right\|_2^2 = \left\| \mb{A} - \hat{\mb{A}} \right\|_2^2,
\end{align} 

\vspace{-15pt}
\noindent where matrix $\hat{\mb{A}}$ involves the reconstructed feature vectors of nodes in the network. 

Meanwhile, different from representation learning of independent data instance, the nodes in networks are strongly correlated. For the nodes with connections, they should have closer representations while those without connections should have relatively different representations instead. Such an intuition can be formally represented as the loss term: \vspace{-6pt}
\begin{equation}\label{equ:Lc}
\mathcal{L}_c(g) = \sum_{v_i, v_i \in \mathcal{V}_g} s(i,j) \left\| \mb{z}_i - \mb{z}_j \right\|_2^2 = \mbox{Tr}(\mb{Z}^\top \mb{L} \mb{Z}),
\end{equation}

\vspace{-12pt}
\noindent where $s(i,j) = +1$ if $(v_i, v_j) \in \mathcal{E}_g$ and $s(i,j) = -1$ if $(v_i, v_j) \notin \mathcal{E}_g$. Matrix $\mb{Z} \in \mathbb{R}^{|\mathcal{V}_g| \times d}$ contains the representation feature vectors of nodes in network $g$, and $\mb{L} = \mb{D} - \mb{S}$ denotes the Laplacian matrix of $\mb{S}$ (where $\mb{S}$ contains entry $S(i,j) = s(i,j)$ and $\mb{D}$ is the diagonal matrix of $\mb{S}$).

Formally, the joint objective functions of the unit model $M^{(j)}_i$ on batch $\mathcal{T}^{(j)}_i$ can be represented as  \vspace{-6pt}
\begin{equation}
\min_{\mb{\theta}} \sum_{g \in \mathcal{T}^{(j)}_i} \Big(\mathcal{L}_e(g) +  \mathcal{L}_c(g) \Big) + \alpha \cdot \mathcal{L}_{reg}(\mb{\theta}^{(j)}_i),
\end{equation}

\vspace{-10pt}
\noindent where $\alpha$ is the weight and $\mathcal{L}_{reg} (\mb{\theta})$ denotes the regularization (i.e., sum of $L_2$ norms) on the variables $\mb{\theta}^{(j)}_i=(\{\mb{W}^k\}_k, \{\mb{b}^k\}_k, \{\hat{\mb{W}}^k\}_k, \{\hat{\mb{b}}^k\}_k)$ involved in model $M^{(j)}_i$.

\vspace{-10pt}
\subsection{Generation Validation and Parent Model Selection}
\vspace{-8pt}

Before the model generation evolution, each unit model will be evaluated to get its fitness scores. Formally, when evaluating the performance of unit models in generation $\mathcal{G}^{(j)}$, a shared validation set $\mathcal{V}^{(j)}$ will be sampled from the pool. For each trained unit model $M^{(j)}_i \in \mathcal{G}^{(j)}$, its introduced error on $\mathcal{V}^{(j)}$ can be represented as  \vspace{-6pt}
\begin{equation}\mathcal{L}(M^{(j)}_i | \mathcal{V}^{(j)}) = \sum_{g \in \mathcal{V}^{(j)}} \left(\mathcal{L}_e(g | M^{(j)}_i) +  \mathcal{L}_c(g | M^{(j)}_i) \right).\end{equation}

\vspace{-10pt}
In validation, some unit models can introduce errors with negative values, which will make $\exp(- {\mathcal{L}}(M^{(j)}_i| \mathcal{V}^{(j)}) )$ approach $\infty$ in applications. Due to this reason, the error terms of all the models will be normalized into a list $\Big(\bar{\mathcal{L}}(M^{(j)}_1| \mathcal{V}^{(j)}), \bar{\mathcal{L}}(M^{(j)}_2| \mathcal{V}^{(j)}), \cdots, \bar{\mathcal{L}}(M^{(j)}_m| \mathcal{V}^{(j)}) \Big)$, where the min-max normalization can be adopted and we have $\bar{\mathcal{L}}(M^{(j)}_i | \mathcal{V}^{(j)}) \in [0, 1]$. In {\our}, unit models which perform better on the validation set will have a higher chance to be selected for generating the child models. Formally, with the normalized errors, the unit model selection probabilities in generation $\mathcal{G}^{(j)}$ can be denoted as $p^{(j)}_1, p^{(j)}_2, \cdots, p^{(j)}_m$, where  \vspace{-5pt}
\begin{equation}
p^{(j)}_i = \frac{\exp(- \bar{\mathcal{L}}(M^{(j)}_i | \mathcal{V}^{(j)}))}{\sum_{M^{(j)}_i \in \mathcal{G}^{(j)}} \exp(- \bar{\mathcal{L}}(M^{(j)}_i | \mathcal{V}^{(j)}))}.
\end{equation}

 \vspace{-8pt}
\noindent From generation $\mathcal{G}^{(j)}$, based on the above probability definition, $m$ pairs of unit models $\big((M^{(j)}_i, M^{(j)}_k)_1$, $(M^{(j)}_p, M^{(j)}_q)_2$, $\cdots, (M^{(j)}_r, M^{(j)}_s)_m \big)$ will be selected (with replacement) as the parent models for crossover to be introduced in the next subsection.

\vspace{-10pt}
\subsection{Genetic Evolution: Crossover}
\vspace{-8pt}

The crossover operation denotes the process of mixing the gene (i.e., variables) of parent models for generating the child models. In {\our}, we will use bi-crossover, i.e., crossover from two parent models to generate one child model. Between the parent models, they will also compete to pass their variables to their child model, where ``good'' parent model will have a large chance. 

Formally, based on the parent models $(M^{(j)}_i, M^{(j)}_k)$, we can represent their generated child model as $M^{(j+1)}_l$, whose variables can be denoted as $\bar{\mb{\theta}}^{(j+1)}_l$. In crossover, for each entry in child model variable $\bar{\mb{\theta}}^{(j+1)}_l$, parent model $M^{(j)}_i$ wins the entry (i.e., assign it with value from $\mb{\theta}^{(j)}_i$) with a chance $\frac{p^{(j)}_i}{p^{(j)}_i+p^{(j)}_k}$ and model $M^{(j)}_i$ wins it with a chance $\frac{p^{(j)}_k}{p^{(j)}_i+p^{(j)}_k}$. 

\vspace{-10pt}
\subsection{Genetic Evolution: Mutation}
\vspace{-8pt}

Via crossover, we can represent the variables of the generated child models as $\{\bar{\mb{\theta}}^{(j+1)}_1, \bar{\mb{\theta}}^{(j+1)}_2, \cdots, \bar{\mb{\theta}}^{(j+1)}_m\}$. In the learning process of {\our}, these newly evolved unit model have a certain chance to mutate, which denotes the unexpected changes in the variable values. From the model learning perspective, the model variable mutation provides the opportunity to jump out from the local minima and approach the global minimum. For the bad mutations, the model may get stuck into another worse local minimum instead, which will have less opportunities to evolve the child models in the next generations.

Formally, given the unit model variable $\bar{\mb{\theta}}^{(j+1)}_i$ of model $M^{(j+1)}_i \in \mathcal{G}^{(j+1)}$, we can denote the model variable after mutation as ${\mb{\theta}}^{(j+1)}_1$, where variable vector entry  \vspace{-6pt}
\begin{equation}{\mb{\theta}}^{(j+1)}_i(k) =
\begin{cases}
rand(0, 1) & \mbox{, if } \bar{p} \le \hat{p}, \\
\bar{\mb{\theta}}^{(j+1)}_i (l) & \mbox{, otherwise},
\end{cases}\end{equation}

 \vspace{-8pt}
\noindent where $\hat{p}$ denotes the mutation probability and $\bar{p} = rand(0,1)$ is a random number. In the case that $\bar{p} \le \hat{p}$, entry ${\mb{\theta}}^{(j+1)}_i(k)$ will be mutated with a random variable value $rand(0, 1)$.

\vspace{-10pt}
\subsection{Output Ensemble}
\vspace{-8pt}

Via iterative model learning, we can represent the final generation of unit models as $\mathcal{G}^{(K)} = \{M^{(K)}_1, M^{(K)}_2, \cdots, M^{(K)}_m\}$. In the output result ensemble step, we will apply all the unit models in $\mathcal{G}^{(K)}$ to the sampled pool set $\mathcal{P}$, where the representation of node $v_i$ learned by unit model $M^{(K)}_j$ on network $g \in \mathcal{P}$ can be represented as $\mb{z}_{i,j}(g)$. In the case that $v_i$ doesn't appear in network $g$, random padding will be adopted to generate its representation vector $\mb{z}_{i,j}(g)$. Different unit models can capture different properties of nodes in their representations. The final output representation feature vector of node $v_i \in \mathcal{V}$ can be defined as the concatenation of learned representation from the unit models. Formally, the output result of node $v_i$ can be defined as  \vspace{-6pt}
\begin{equation}
\mb{z}_{i} = \mb{z}_{i,1} \oplus  \mb{z}_{i,2} \oplus \cdots \oplus  \mb{z}_{i,m},
\end{equation}

\vspace{-9pt}
\noindent where operator $\oplus$ concatenates the vectors and $\mb{z}_{i,1} = [\mb{z}_{i,j}(g_1), \mb{z}_{i,j}(g_2), \cdots, \mb{z}_{i,j}(g_{|\mathcal{P}|})]$.

\vspace{-10pt}
\section{Model Analysis}\label{sec:analysis}
\vspace{-10pt}

In this section, we will analyze {\our} from its performance, running time and space cost, which will illustrate the advantages of {\our} compared with other existing deep learning models.

\vspace{-10pt}
\subsection{Performance Analysis}
\vspace{-8pt}

Model {\our}, in a certain sense, can also be called a ``deep'' model. Instead of stacking multiple hidden layers inside one single model like existing deep learning models, {\our} is deep since the unit models in successive generations are generated by a namely ``evolution layer'' which performs the \textit{validation}, \textit{crossover}, \textit{mutation} and \textit{selection} operations connecting generations. Between the generations, these ``evolution operations'' mainly work on the unit model variables, which allows the immigration of learned knowledge from generation to generation. In addition, via these generations, the last generation in {\our} can also capture the overall patterns of the dataset. Since the unit models in different generation are built with different sampled training batches, as more generations are involved, the dataset will be samples thoroughly for learning {\our}. There have been lots of research works done on analyzing the convergence, performance bounds of genetic algorithms \cite{R94}, which can provide the theoretic performance foundations for {\our}. Due to the differences in parent model selection, crossover, mutation operations and different sampled training batches, the unit models in the generations of {\our} may perform quite differently. With the diverse result combined from these different learning models, {\our} is able to achieve better performance than each of the unit models, which have been effectively demonstrated in the ensemble learning research works \cite{ZWT02}. 

\begin{table*}[t]
\vspace{-25pt}
\caption{Network Recovery Task Results. PS1: the default parameter setting; PS2: sub-network size: 30, pool size: 400, batch size: 35, generation unit model number: 20, generation number: 20.}
\label{tab:result}
\scriptsize
\centering
\begin{tabular}{c  ccc  ccc  c}
\toprule
&\multicolumn{3}{c}{Network Recovery (AUC)} &\multicolumn{3}{c}{Network Recovery (Prec@500)} & {Learning Time Cost}\\
\cmidrule(lr){2-4}\cmidrule(lr){5-7}\cmidrule(lr){8-8}
np-ratio &{1} &{5} & 10 &{1} &{5} & 10	& (in seconds) \\
\midrule 
{{\our}(PS2)}	&\textbf{0.793} {\footnotesize {\color{blue} (1)}} & \textbf{0.793} {\footnotesize {\color{blue} (1)}} & \textbf{0.792} {\footnotesize {\color{blue} (1)}} 	&\textbf{0.955} {\footnotesize {\color{blue} (1)}} &\textbf{0.822} {\footnotesize {\color{blue} (1)}} &\textbf{0.652} {\footnotesize {\color{blue} (1)}}	& \textbf{169.042} {\footnotesize {\color{blue} (2)}} \\
\midrule 
{{\our}(PS1)}	&\textbf{0.663} {\footnotesize {\color{blue} (3)}} &\textbf{0.663} {\footnotesize {\color{blue} (3)}} &\textbf{0.662} {\footnotesize {\color{blue} (3)}}	&\textbf{0.652} {\footnotesize {\color{blue} (2)}} &\textbf{0.146} {\footnotesize {\color{blue} (3)}} &0.046 {\footnotesize {\color{blue} (4)}}	&\textbf{72.117} {\footnotesize {\color{blue} (1)}} \\
\midrule 
{\linemodel} \cite{TQWZYM15}	&0.254 {\footnotesize {\color{blue} (6)}} &0.254 {\footnotesize {\color{blue} (6)}} &0.253 {\footnotesize {\color{blue} (6)}}	&0.106 {\footnotesize {\color{blue} (6)}} &0.018 {\footnotesize {\color{blue} (5)}}	&0.006 {\footnotesize {\color{blue} (5)}} 	& \textbf{252.319} {\footnotesize {\color{blue} (3)}} \\
\midrule 
{\deepwalk} \cite{PAS14}	&0.533 {\footnotesize {\color{blue} (5)}} &0.531 {\footnotesize {\color{blue} (5)}} &0.532 {\footnotesize {\color{blue} (5)}}	&0.524 {\footnotesize {\color{blue} (5)}} &\textbf{0.146} {\footnotesize {\color{blue} (3)}} &\textbf{0.070} {\footnotesize {\color{blue} (3)}}	&460.624 {\footnotesize {\color{blue} (5)}} \\
\midrule 
{\nodevec} \cite{GL16}	&\textbf{0.704} {\footnotesize {\color{blue} (2)}} &\textbf{0.703} {\footnotesize {\color{blue} (2)}} &\textbf{0.704} {\footnotesize {\color{blue} (2)}}	&0.528 {\footnotesize {\color{blue} (4)}} &0.012 {\footnotesize {\color{blue} (6)}} &0.000 {\footnotesize {\color{blue} (6)}}	&734.996 {\footnotesize {\color{blue} (6)}} \\
\midrule 
{\hpe} \cite{CTLY16}		&0.593 {\footnotesize {\color{blue} (4)}} &0.595 {\footnotesize {\color{blue} (4)}} &0.594 {\footnotesize {\color{blue} (4)}}	&\textbf{0.534} {\footnotesize {\color{blue} (3)}} &\textbf{0.186} {\footnotesize {\color{blue} (2)}} &\textbf{0.094} {\footnotesize {\color{blue} (2)}}	&2927.118 {\footnotesize {\color{blue} (4)}} \\
\bottomrule
\end{tabular}
\vspace{-15pt}
\end{table*}

\vspace{-10pt}
\subsection{Space and Time Complexity Analysis}
\vspace{-8pt}

According the the model descriptions provided in Section~\ref{sec:method}, we summarize the parameters used in {\our} as follows, which will help analyze its space and time complexity.

\vspace{-2pt}

\begin{itemize}

\item \textit{Dataset Sampling}: Original data size: $n$. Sub-instance size: $n'$. Pool size: $p$.
\vspace{-2pt}
\item \textit{Model Learning}: Generation number: $K$. Generation size: $m$. Feature vector size: $d$. Training/validation batch size: $b$.

\end{itemize}

\vspace{-8pt}

\noindent \textbf{Space Complexity}: Given a large-scale network with $n$ nodes, the space cost required for storing the whole network in a matrix representation is $O(n^2)$. Meanwhile, via network sampling, we can obtain a pool of sub-networks, and the space required for storing these sub-networks takes $O\left(p (n')^2 \right)$. Generally, in application of {\our}, $n'$ can take a very small number, e.g., $50$, and $p$ can take value $p = c \cdot \frac{n}{n'}$ ($c$ is a constant) so as to cover all the nodes in the network. In such a case, the space cost of {\our} will be linear to $n$, i.e., $O(cn' n)$, which is much smaller than $O(n^2)$.

\noindent \textbf{Time Complexity}: Depending on the specific unit models used in composing {\our}, we can represent the introduced time complexity of learning one unit model on the original network with $n$ nodes as $O(f(n))$, where $f(n)$ is usually a high-order function of $n$. Meanwhile, for learning {\our} on the sampled sub-networks with $n'$ nodes, all the introduced time cost will be $O\left(Km (b\cdot f(n') + d \cdot n') \right)$, where term $d\cdot n'$ (an approximation of the variable number) represents the cost introduced in the unit model crossover and mutation about the model variables. Here, by assigning $b$ with a fixed value $b = c \cdot \frac{n}{n'}$, the time complexity of {\our} will be reduced to $O\left(Kmc\frac{f(n')}{n'} \cdot n + Kmd n' \right)$, which is linear to $n$.

\vspace{-12pt}
\subsection{Advantages Over Deep Learning Models}
\vspace{-8pt}

Compared with existing deep learning models based on the whole dataset, the advantages of {\our} are summarized below:
\begin{itemize}
\vspace{-8pt}
\item \textit{Less Data for Unit Model Learning}: For each unit model, which are of a ``shallow'' and ``narrow'' structure (shallow: less or even no hidden layers, narrow: based on sampled sub-instances with much smaller dimensions), which result in far less variables and less data consumption for learning each unit model.
\vspace{-2pt}
\item \textit{Less Computational Resources}: Each unit model is of a much simpler structure, learning process of which consumes far less computational resources. Detailed analysis about the running time and space cost will be provided in the following subsections.
\vspace{-2pt}
\item \textit{Less Parameter Tuning}: {\our} accepts both deep and shallow learning models as the unit learning in the model building, and the hyper-parameters can also be shared among the unit models, which lead to far less hyper-parameters to tune in the learning process.

\vspace{-2pt}

\item \textit{Sound Theoretic Explanation}: The unit learning model, genetic algorithm and ensemble learning (aforementioned) can all provide the theoretic foundation for {\our}, which will lead to sound theoretic explanation of both the learning result and the {\our} model itself.
\end{itemize}

\vspace{-5pt}


\vspace{-10pt}
\section{Experiments}\label{sec:experiment}
\vspace{-10pt}

To test the effectiveness of the proposed model, in this part, extensive experiments will be done to compare {\our} with existing state-or-the-art baseline methods.

\vspace{-10pt}
\subsection{Network Structured Data Experimental Settings}
\vspace{-8pt}

The network dataset used in the experiments is crawled from Twitter, which involves $5,120$ users and $130,576$ social connections among the user nodes. In the experiments, except the parameters to be analyzed or those with specified values, the other parameters are set with the following default values: sub-network size: 10, pool size: 200, batch size: 10, generation unit model number: 10, generation number: 20, mutation probability: 0.01, np-ratio: 1.

The network representation learning comparison models used in this paper are listed as follows
\begin{itemize}
\vspace{-3pt}
\item \textit{{\our}}: Model {\our} proposed in this paper is based on the genetic algorithm and ensemble learning, which effectively learns the network representations based on several generations of small-sized unit models instead.
\vspace{-3pt}
\item \textit{{\linemodel}}: The {\linemodel} model is a scalable network embedding model proposed in \cite{TQWZYM15}, which preserves both the local and global network structures. 
\vspace{-3pt}
\item \textit{{\deepwalk}}: The {\deepwalk} model \cite{PAS14} extends the word2vec model \cite{MSCCD13} to the network embedding scenario. {\deepwalk} uses local information obtained from truncated random walks to learn latent representations.
\vspace{-3pt}
\item \textit{{\nodevec}}: The {\nodevec} model \cite{GL16} introduces a flexible notion of a node's network neighborhood and design a biased random walk procedure to sample the neighbors for node representation learning.
\vspace{-3pt}
\item \textit{{\hpe}}: The {\hpe} model \cite{CTLY16} is originally proposed for learning user preference in recommendation problems, which can effectively project the information from heterogeneous networks to a low-dimensional space.

\end{itemize}

\vspace{-8pt}
In this paper, we propose to use the network recovery task to evaluate the learned representation features from the comparison methods. Based on the complete input network structure, we can represent the existing connections as the positive set. Meanwhile, to control the ratio of negative/positive instances (i.e., class imbalanced), a subset of the non-existing connections are randomly sampled as the negative instance, whose number is controlled by the np-ratios in $\{1, 5, 10\}$. Based on the learned representation feature vectors, we aim at inferring the existence of these links. Furthermore, the network recovery results are evaluated by metrics, like AUC and Precision@K effectively. The learning efficiency of the baseline methods are evaluated by counting their running time.

\begin{figure}[t]
\vspace{-25pt}
\begin{minipage}[b]{0.33\linewidth}
      \centering
      \includegraphics[width=1.0\textwidth]{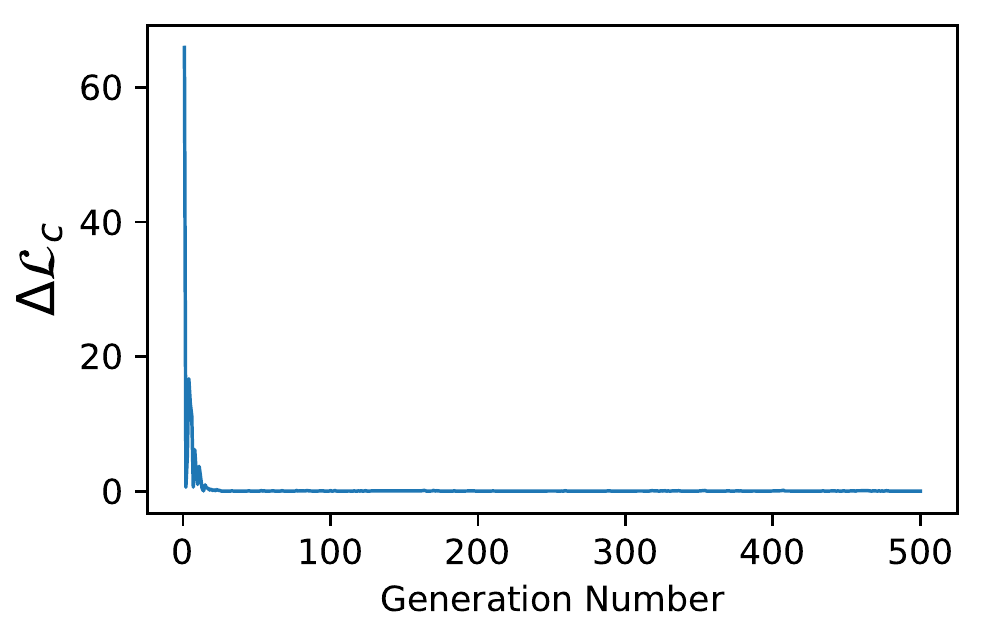}
\caption{Convergence Analysis.}\label{fig:convergence}
\end{minipage}
\begin{minipage}[b]{0.66\linewidth}
\centering
\subfigure[Network Rec: AUC]{\label{fig:Sampling_Parameter_LP_AUC}
    \begin{minipage}[l]{0.45\columnwidth}
      \centering
      \includegraphics[width=1.0\textwidth]{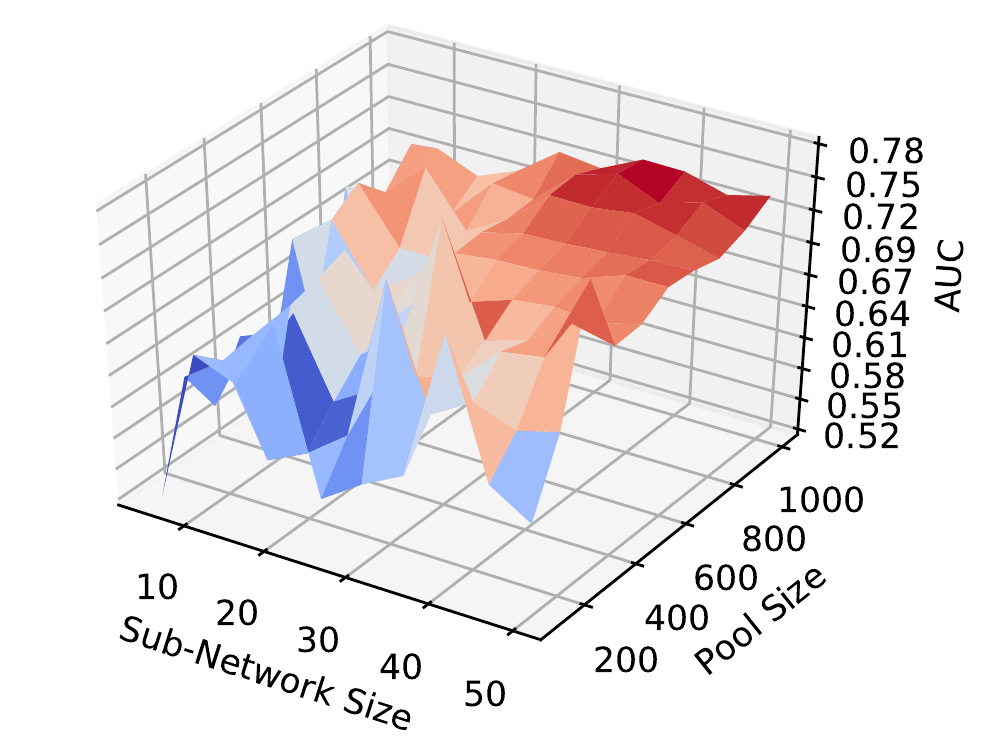}
    \end{minipage}
}
\subfigure[Network Rec: Prec@500]{\label{fig:Sampling_Parameter_LP_Prec}
    \begin{minipage}[l]{0.45\columnwidth}
      \centering
      \includegraphics[width=1.0\textwidth]{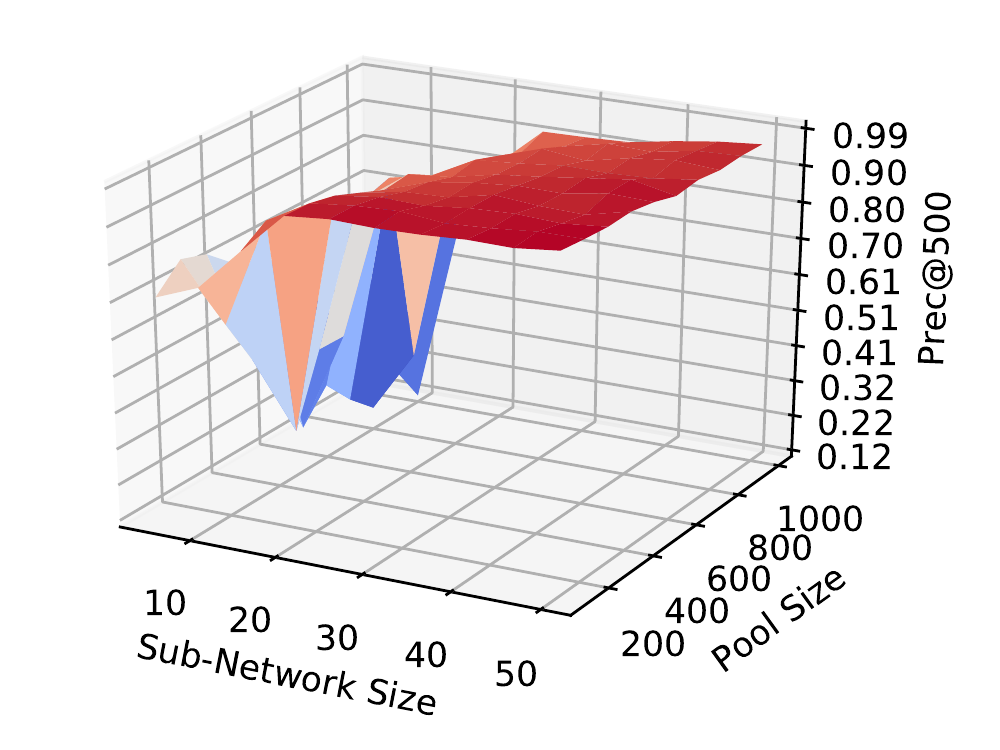}
    \end{minipage}
}
\vspace{-10pt}
\caption{Analysis of Sampling Parameters.}\label{fig:Sampling_Parameter}
\end{minipage}
\end{figure}

\begin{figure}[t]
\begin{minipage}[b]{0.5\linewidth}
\centering
\subfigure[AUC]{ \label{fig:generation_size_LP}
    \begin{minipage}[l]{0.45\columnwidth}
      \centering
      \includegraphics[width=1.0\textwidth]{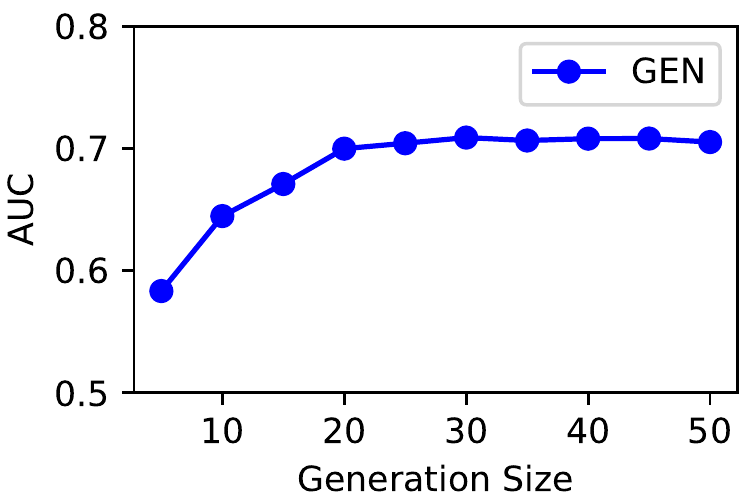}
    \end{minipage}
}
\subfigure[Prec@500]{ \label{fig:generation_size_LP_prec}
    \begin{minipage}[l]{0.45\columnwidth}
      \centering
      \includegraphics[width=1.0\textwidth]{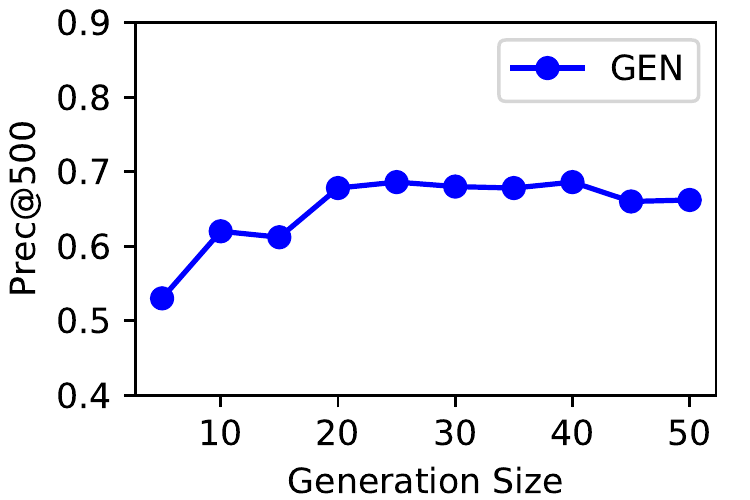}
    \end{minipage}
}
\vspace{-5pt}
\caption{Analysis of Model Generation Size.}\label{fig:generation_size}\vspace{-10pt}
\end{minipage}
\begin{minipage}[b]{0.5\linewidth}
\centering
\subfigure[AUC]{ \label{fig:batch_size_LP}
    \begin{minipage}[l]{0.45\columnwidth}
      \centering
      \includegraphics[width=1.0\textwidth]{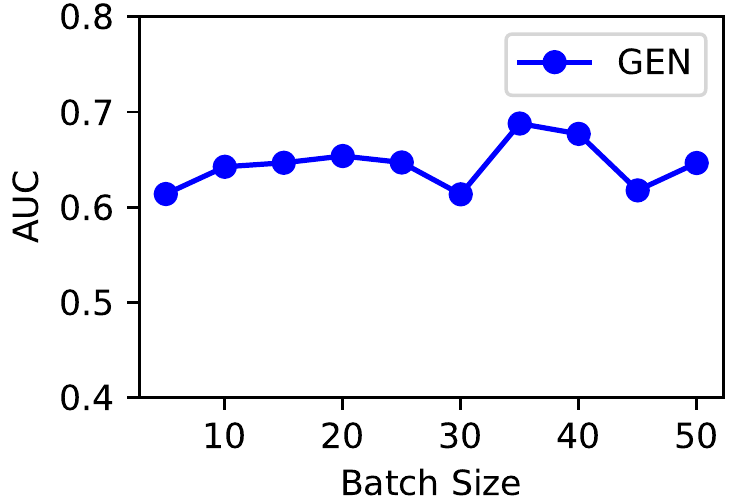}
    \end{minipage}
}
\subfigure[Prec@500]{ \label{fig:batch_size_LP_prec}
    \begin{minipage}[l]{0.45\columnwidth}
      \centering
      \includegraphics[width=1.0\textwidth]{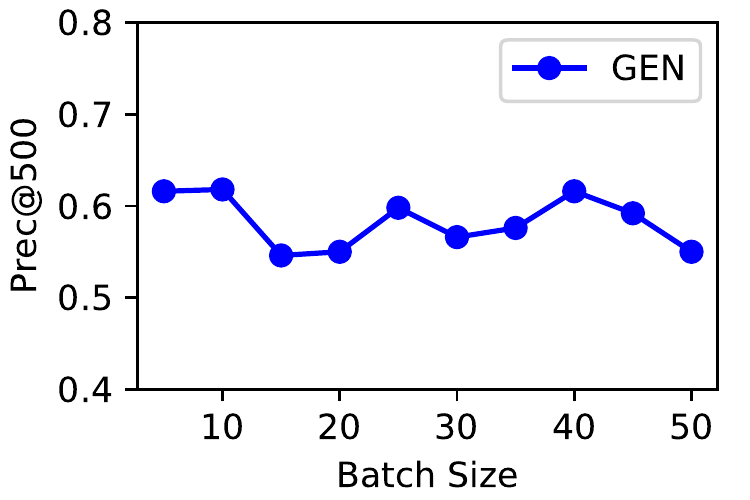}
    \end{minipage}
}
\vspace{-5pt}
\caption{Analysis of Model Batch Size.}\label{fig:batch_size}\vspace{-10pt}
\end{minipage}
\end{figure}

\vspace{-10pt}
\subsection{Network Data Experimental Results, Convergence Analysis and Setting Analysis}\label{subsec:convergence}
\vspace{-8pt}

In Figure~\ref{fig:convergence}, we show the changes of loss term $\mathcal{L}_c$ introduced in each generation by model {\our} on the validation set. In the figure, the x axis denotes the generation count, and the y axis denotes the changes of loss term $\mathcal{L}_c(\cdot)$ compared with the previous generation introduced by all the unit models on all the graphs in the validation set. According to the plot, model {\our} can converge very quickly with less than $20$ generations. Therefore, we use $K=20$ as the default value for the generation parameter in the experiments.

In Table~\ref{tab:result}, we show the comparison of experimental results of network recovery based on the learning features obtained by {\our} and other baseline methods. The performance rank of these methods is indicated by the blue numbers in the table, and the top $3$ results are in bolded font. According to Table~\ref{tab:result}, model {\our} with PS2 will introduce a much better performance in both effectiveness and efficiency than the baseline methods. The time cost of {our} (PS2) ranks \#2 and loses to {our} (PS1) only, as {our} (PS1) involves much less unit models, smaller sub-networks and pools. 

The sampled sub-network pool is the foundation of {\our}, whose parameter selection may affect the performance of {\our} a lot. Two strongly correlated parameters, i.e., sub-network size and pool size, are involved in the sampling. In the analysis, we change the sub-network size with values in $\{5, 10, 15, \cdots, 50\}$ and pool size with values in $\{100, 200, \cdots, 1000\}$, and the performance by {\our} evaluated by AUC and Prec@500 are illustrated in Figure~\ref{fig:Sampling_Parameter}. According to the plot, a large pool of large-sized sub-networks will always lead to better performance. Meanwhile, as shown in Figure~\ref{fig:Sampling_Parameter_LP_Prec}, we also observe that for the sub-networks greater than a certain size (e.g., 20), the impact of the pool size on {\our} becomes minor. The potential reason can be a certain number of such kinds of sub-networks can already capture the network structure already, and further increase the pool size will not introduce significant improvement.

As shown in Figure~\ref{fig:generation_size}, we show the sensitivity analysis of the generation size (i.e., unit model population in each generation). According to the results, as the generation size increases with values in $\{5, 10, 15, \cdots, 50\}$, the performance of the {\our} model increase steadily. This is easy to understand, as more unit models are involved in the generations, {\our} will have more options to select for evolution. Meanwhile, according to the plots in Figures~\ref{fig:generation_size_LP}-\ref{fig:generation_size_LP_prec}, as the generation size further increases from $20$ to $50$, the improvement is very minor, especially for AUC, and it seems $20$ unit models is already a good population size for the input dataset.

In Figure~\ref{fig:batch_size}, we show the sensitivity analysis of the batch size (i.e., sub-network number in training/validation batches), which changes with values in range $\{5, 10, 15, \cdots, 50\}$. As shown in the figure, with the increase of batch size parameter, there exist some fluctuations in the performance of {\our}. It seems a small batch can achieve very close performance with the large batches. Among all these batch size values in Figure~\ref{fig:batch_size}, values $35$/$40$ can achieve the best performance among all the parameters for both of these two tasks.

\vspace{-10pt}
\subsection{Experimental Results with Other Data Sets and Unit Models}\label{subsec:other_results}
\vspace{-8pt}

\begin{table}[t]
\vspace{-20pt}
\begin{minipage}[b]{0.45\linewidth}
\caption{Experiments on MNIST Dataset.}\label{tab:mnist_result}
\centering    
\scriptsize
 \begin{tabular}{| l | c |} 
 \hline
 Comparison Methods & Accuracy Rate\% \\
  \hline\hline
{\our} (CNN) &99.37   \\ 
 \hline
 LeNet-5 &99.05 \cite{726791} \\
  \hline
 gcForest &99.26 \cite{ijcai2017-497} \\
  \hline
 Deep Belief Net &98.75 \cite{HOT06} \\
  \hline
 Random Forest &96.8 \cite{ijcai2017-497} \\
  \hline
 SVM (rbf) &98.60 \cite{DS02} \\
 \hline
\end{tabular}
\end{minipage}\hfill   
\begin{minipage}[b]{0.5\linewidth}
\caption{Experiments on Other Datasets.}\label{tab:other_result}
\scriptsize
\centering
 \begin{tabular}{| l | c | c | c |} 
 \hline
 \multirow{2}{*}{Comparison Methods} &	\multicolumn{3}{c|}{Accuracy Rate \% on Datasets} \\
 \cline{2-4}
  & YEAST & ADULT & LETTER \\
  \hline \hline
 {\our} (MLP) &63.70		&87.05		&96.90		\\
  \hline
 MLP &62.05 	&85.03		&96.70		\\
  \hline
 gcForest & 63.45		& 86.40		& 97.40		\\
   \hline
 Random Forest & 60.44		& 85.63		& 96.28		\\
  \hline
 SVM (rbf) & 40.76		& 76.41		&97.06		\\
  \hline
 kNN (k=3) & 48.80		& 76.00		&95.23		\\
 \hline
\end{tabular}
\end{minipage}
\vspace{-15pt}
\end{table}

Besides the extended autoencoder model and the network datasets, we have also test the effectiveness of {\our} on other datasets and with other unit models. In Table~\ref{tab:mnist_result}, we show the experimental results of {\our} and other baseline methods on the MNIST hand-written image datasets. The dataset contains $60,000$ training instances and $10,000$ testing instances, where each instance is a $28 \times 28$ image with labels denoting their corresponding numbers. CNN is used as the unit model in {\our}, which involves 2 convolutional layers, 2 max-pooling layers, and two fully connection layers (with a $0.2$ dropout rate). Meanwhile, in Table~\ref{tab:other_result}, we provide the learning results on three other datasets, including YEAST\footnote{https://archive.ics.uci.edu/ml/datasets/Yeast}, ADULT\footnote{https://archive.ics.uci.edu/ml/datasets/adult} and LETTER\footnote{https://archive.ics.uci.edu/ml/datasets/letter+recognition}. MLP is used as the unit model in {\our} for these three datasets. For the ensemble strategy in these experiments, the best unit model is selected to generate the final prediction output. According to the results, compared with the baseline methods, {\our} can also perform very well with CNN and MLP on the image and other categories of datasets.

\vspace{-10pt}
\section{Conclusion}\label{sec:conclusion}
\vspace{-8pt}

In this paper, we have introduced an alternative approach to deep learning models, which is called the ``Genetic Evolutionary Network'' ({\our}). {\our} adopts an evolving architecture for unit model learning, where good unit models will be selected to generate the offsprings. {\our} is far more efficient than existing deep learning models. Based on the sampled sub-instances, {\our} can learn unit models with much less variables, computational resources and training data, where the unit models can be either deep models or traditional machine learning models. Furthermore, the theoretic foundation of genetic algorithm and ensemble learning, also helps explain the information inheritation through generations and output integration from the unit model population in {\our}.

\bibliographystyle{abbrv}
\bibliography{reference}

\end{document}